\documentclass[letterpaper]{article} 
\usepackage[preprint]{aaai2027}  
\usepackage[hyphens]{url}  
\usepackage{graphicx} 
\usepackage{natbib}  
\usepackage{caption} 
\usepackage{subcaption}
\usepackage{algorithm}
\usepackage{algorithmic}

\usepackage{newfloat}
\usepackage{listings}
\DeclareCaptionStyle{ruled}{labelfont=normalfont,labelsep=colon,strut=off} 
\floatstyle{ruled}
\newfloat{listing}{tb}{lst}{}
\floatname{listing}{Listing}

\usepackage{booktabs}
\usepackage{amsmath}
\usepackage{amssymb}
\usepackage{multirow}
\usepackage{makecell}
\usepackage{enumitem}
\usepackage{tikz}
\usetikzlibrary{shapes.geometric, arrows.meta, positioning, fit}

\title{EvoOntology: A Self-Evolving Ontology Layer for Data Agents}
\author {
    Meiduo Chong\textsuperscript{\rm 1},
    Shaolei Zhang\textsuperscript{\rm 1}\thanks{Corresponding author: Shaolei Zhang.},
    Ju Fan\textsuperscript{\rm 1},
    Xiaoyong Du\textsuperscript{\rm 1}
}

\affiliations {
    \textsuperscript{\rm 1}Renmin University of China\\
    zhongmeiduo210@ruc.edu.cn,
    zhangshaolei98@ruc.edu.cn
}

\begin{document}

\maketitle

\begin{abstract}
%
Data agents aim to fulfill natural-language instructions over heterogeneous data, including tables, files, and databases. However, data agents face a challenging \emph{agent--data gap}: heterogeneous data resides outside the agent, while the agent can access it (e.g., column names and file paths) only through generic tools. 
Existing approaches either let agents directly explore raw data sources or inject manually constructed semantic layers into prompts. However, neither scales well to large heterogeneous data sources nor adapts to different agent behaviors. 
In this paper, we introduce \emph{EvoOntology}, a self-evolving ontology layer for data agents. EvoOntology encapsulates the ontology as an MCP server comprising a schema layer, a content layer, and a tool layer, enabling agents to actively query and interact with the ontology at runtime. To this end, we introduce a builder agent for autonomous ontology construction and a self-evolution loop that continuously refines the ontology through attribution-guided typed edits that are accepted only after a backbone-conditional paired evaluation.
Experiments on three well-adopted data-agent benchmarks with four LLM backbones demonstrate that EvoOntology consistently outperforms strong baselines and existing semantic-layer approaches, effectively bridging the \emph{agent--data gap} and enabling more effective interaction with heterogeneous data.
\end{abstract}

\begin{links}
    \link{\quad Code}{https://github.com/ruc-datalab/EvoOntology}
\end{links}

\section{Introduction}

Data agents over heterogeneous data~\cite{ddr_bench,insightbench,bird_bench,data_interpreter,data_copilot} aim to solve natural-language tasks over both structured data (e.g., tables and databases) and unstructured data (e.g., documents and files). To accomplish such tasks, an agent must continuously interact with heterogeneous data sources to gather the information required for producing the final answer. Recent advances in tool use for large language models (LLMs)~\cite{react,toolformer,toolllm,gorilla} have enabled agents to directly access and manipulate external data sources, providing the foundation for such data interactions.

\begin{figure}[t]
\centering
\includegraphics[width=\columnwidth]{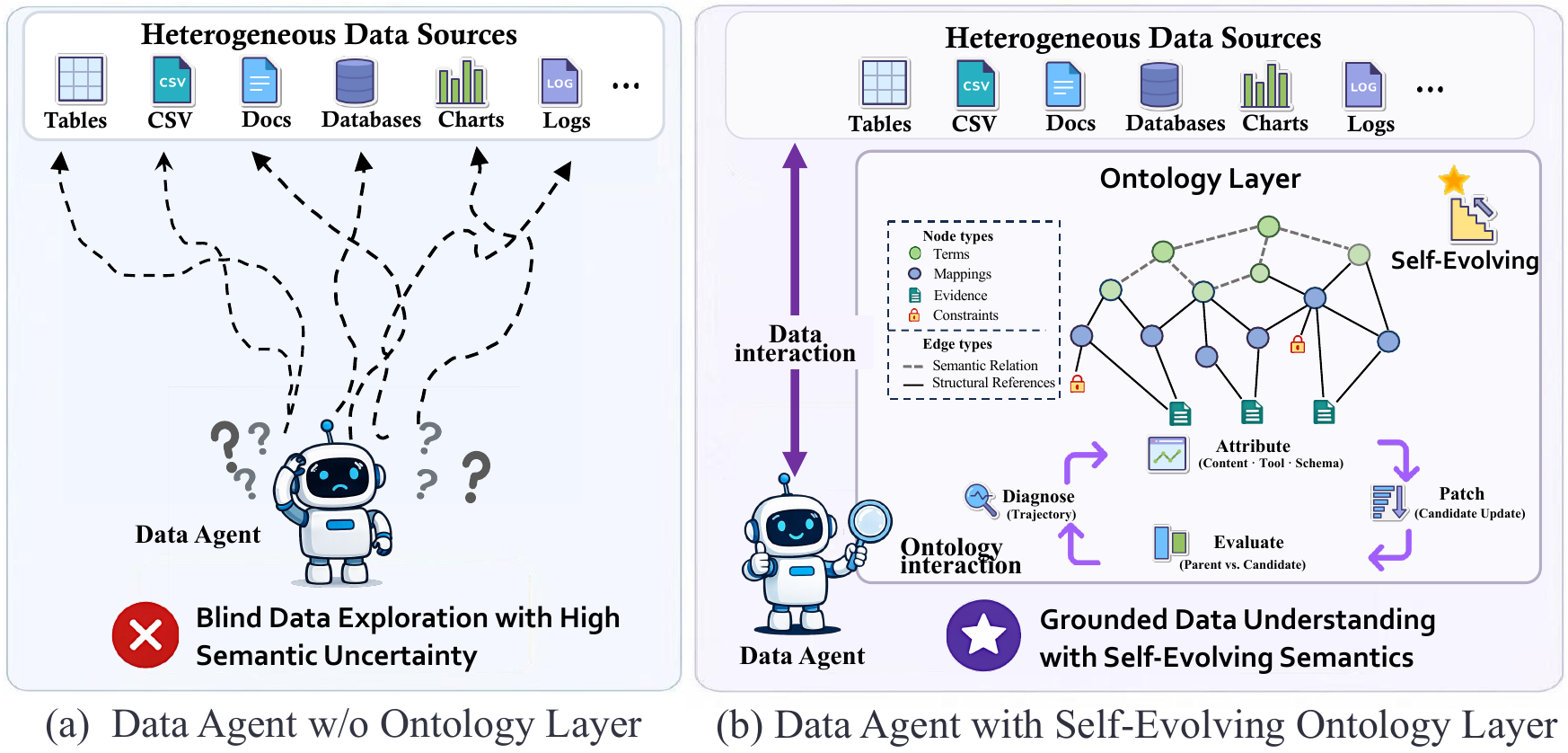}
\caption{A self-evolving ontology layer helps data agents understand heterogeneous data.}
\label{fig:overview}
\end{figure}


However, direct interaction with heterogeneous data raises a fundamental question: \emph{Can a data agent effectively understand heterogeneous data}?
In real-world deployments, data resides outside the agent in the form of relational databases, semi-structured filings, and unstructured documents, while the agent can access the data only through generic tools such as SQL interfaces and file readers. 
A fundamental challenge is that neither the structure nor the content of these heterogeneous data sources is known a priori. As a result, the agent has to blindly explore the underlying data by repeatedly issuing probing queries, guessing where the requested concepts are located, and inspecting potentially irrelevant content. This mismatch creates a persistent \emph{agent--data gap}. Bridging this gap requires an \emph{intermediate ontology layer} that explicitly represents domain concepts, grounds the concepts in the underlying data, and enables agents to interact with data at the semantic level rather than the physical level.


Existing approaches to agent--data interaction can be broadly divided into \emph{raw querying} and \emph{semantic-layer-based interaction}. 
Raw-querying methods~\cite{din_sql,mac_sql,chess} allow agents to directly inspect schemas and issue exploratory queries over the underlying data. While effective for small and relatively simple data sources, they scale poorly to wide and heterogeneous data, where agents can easily become trapped in repetitive and inefficient exploration. 
Semantic-layer approaches~\cite{owl_survey,dbt_metrics,knowledge_cards,schema_prompt}, in contrast, provide metadata, including schemas, entities, metrics, and other domain semantics, to guide the agent. However, incorporating the entire semantic layer into the agent context is impractical for large data sources due to context-length limitations. Moreover, existing semantic layers are typically predefined and maintained manually, making them costly to construct and difficult to adapt to new data sources, tasks, and agents. These limitations highlight the need for an effective and scalable ontology intermediate layer to bridge the \emph{agent--data gap}.

In this paper, we advance the intermediate layer between agents and data from static semantic descriptions to an \emph{interactive ontology layer} that agents can flexibly access through tools. Autonomously constructing such an ontology is inherently challenging because both data sources and agent behaviors are diverse and dynamic, requiring the ontology to adapt to both. 
To address this challenge, we introduce \emph{EvoOntology}, a self-evolving ontology layer that continuously adapts to the underlying data and the agents that use it. 
As illustrated in Figure~\ref{fig:overview}, the ontology consists of three components: a \emph{schema layer}, which defines object types and reference rules; a \emph{content layer}, which stores domain knowledge and data mappings; and a \emph{tool layer}, which exposes executable interfaces for agents to access and manipulate the ontology. These components are encapsulated as a Model Context Protocol (MCP) server, enabling agents to actively query and interact with the ontology rather than passively consuming it as contextual metadata.

Specifically, EvoOntology first employs a {builder agent} to construct an initial ontology by issuing probe queries over the underlying data sources and grounding each ontology entry in the observed data. EvoOntology then continuously refines the ontology based on agent interaction trajectories. Specifically, it performs attribution analysis to identify deficiencies in the current ontology, proposes targeted refinements to its schema, content, or tools, and accepts each refinement only after it passes a paired evaluation on a held-out validation set. Through this iterative self-evolution process, the ontology continuously adapts to both heterogeneous data and agent behaviors, progressively bridging the \emph{agent--data gap}.

%
%

In summary, our main contributions are as follows:
\begin{itemize}[leftmargin=1em, itemsep=0ex, topsep=0ex]
    \item \textbf{Interactive Ontology Layer.} We propose the first autonomous interactive ontology layer for data agents and encapsulate it as an MCP server, enabling agents to query and interact with heterogeneous data through tools.
    \item \textbf{Self-Evolving Ontology.} We introduce a builder agent for autonomous ontology construction and a self-evolving framework that refines the ontology through attribution analysis, targeted refinement, and paired evaluation.
    \item \textbf{Strong Performance.} Extensive experiments on three well-adopted data-agent benchmarks with four LLM backbones demonstrate that EvoOntology consistently and substantially outperforms strong baselines and existing semantic-layer approaches.
\end{itemize}

\section{Related Work}

\textbf{Data Agents on Heterogeneous Data.}\quad Deploying LLMs as data agents is an important step toward automated analytics. Existing approaches fall into two families: \emph{raw querying} and \emph{semantic-layer-based interaction}. Raw-querying agents equip LLMs with schema-reading, query-executing, and file-inspecting tools, exemplified by text-to-SQL agents that generate queries over relational databases~\cite{bird_bench,spider,codeS}, table-QA agents that reason over spreadsheets and web tables~\cite{tabfact,wikitq}, and code-executing analysts that answer business-intelligence questions on CSV files~\cite{insightbench,ds_agent}. Pipeline-style variants organize these tool calls through decomposition, retrieval, and verification~\cite{din_sql,mac_sql,chess,rsl_sql,e_sql,pet_sql}, improving standardized benchmarks while leaving the underlying representation gap untouched. This gap is amplified in heterogeneous settings, where a task may span databases, spreadsheets, and files with different naming conventions, schemas, and granularities. Grounding discovered in one trajectory is typically discarded rather than retained for later tasks. EvoOntology instead amortizes schema discovery across the workload through an ontology layer that preserves such grounding and evolves from agent failures.

\textbf{Semantic Layers.}\quad
Ontology and semantic layers have long connected domain concepts with relational data, ranging from OWL ontologies and metric layers~\cite{owl_survey,dbt_metrics} to LLM-oriented semantic representations and prompt-time metadata~\cite{knowledge_cards,schema_prompt}. Related work also uses LLMs to induce schema or metric descriptions~\cite{llm_schema_gen,text2metric} and feedback to refine prompts or retrievers~\cite{ape,dspy,self_rag}. However, existing layers are typically maintained as static prompt-time metadata. Whether manually authored or automatically induced, they are usually detached from downstream trajectories showing how agents use them. Full-context injection scales poorly to large data sources, while coarse updates provide little basis for identifying which semantic entry affected a downstream decision. This makes targeted, workload-driven maintenance difficult as tasks and agent behavior evolve. EvoOntology instead exposes the ontology through an MCP server for selective runtime access and refines individual entries through typed, evidence-grounded edits admitted by paired validation.

\section{Method}
To reduce manual semantic-layer authoring while adapting the layer to agent behavior, we propose \emph{EvoOntology}, an agent-first builder-and-evolver framework. EvoOntology maintains a versioned ontology state comprising content, schema, and tool layers. A builder agent constructs an evidence-grounded initial state from the training workload and raw sources, while an evolution agent refines it from historical trajectories. The design is \emph{agent-first} in that the ontology is built around the workload, accessed through the agent's tool interface, and adapted from its execution history.

\begin{figure*}[t]
    \centering
    \includegraphics[width=\textwidth]{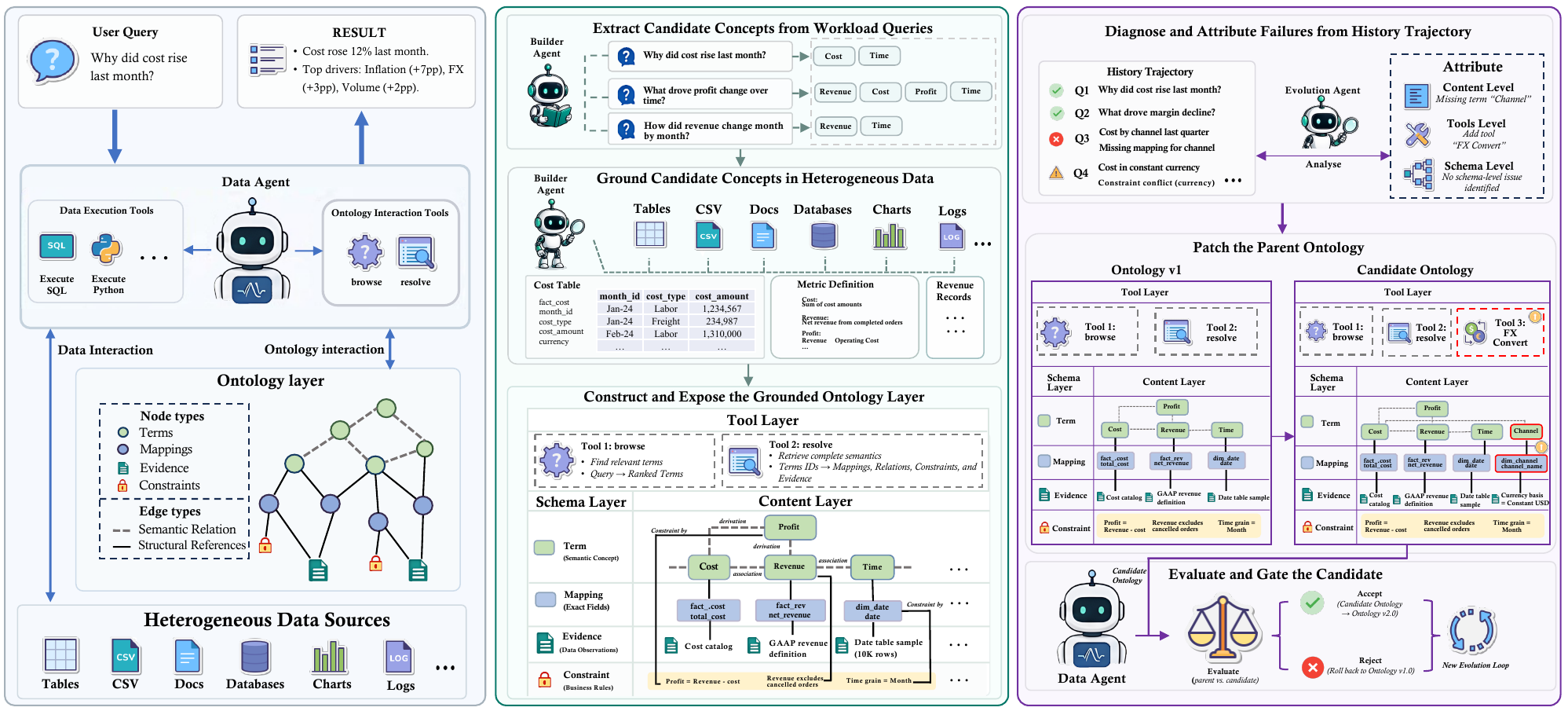}
    \caption{Overview of EvoOntology. It comprises a typed content graph, its object schema, and a runtime tool interface. The builder constructs an evidence-grounded initial state, while the evolution agent refines it from historical interaction trajectories.}
    \label{fig:framework}
\end{figure*}

\subsection{Agent-First Ontology-Layer Architecture}

EvoOntology represents the ontology state at evolution round $t$ as
$\mathcal{L}_t=(\mathcal{S}_t,\Gamma_t,\mathcal{R}_t)$, comprising a \emph{Content Layer} $\mathcal{S}_t$, a \emph{Schema Layer} $\Gamma_t$, and a \emph{Tool Layer} $\mathcal{R}_t$. The three components separate semantic knowledge, its object model, and its runtime exposure. This separation allows the deployed agent to retrieve only the semantics relevant to the current step and allows the evolution agent to update a bounded part of the ontology state.


\textbf{Content Layer.}\quad
The \emph{Content Layer} $\mathcal{S}_t$ is a typed semantic graph with four node families and two edge families. The node families comprise \emph{Terms}, \emph{Mappings}, \emph{Constraints}, and \emph{Evidence}. \emph{Terms} represent domain concepts, \emph{Mappings} ground them to fields and linking paths, \emph{Constraints} govern their valid use, and \emph{Evidence} supports their semantic claims. The edge families comprise \emph{Semantic Relations} and \emph{Structural References}. \emph{Semantic Relations} connect \emph{Terms} through \emph{association}, \emph{hierarchy}, \emph{composition}, \emph{equivalence}, or \emph{derivation}. \emph{Structural References} link \emph{Terms} to \emph{Mappings} and attach \emph{Constraints} and \emph{Evidence} to the objects they govern or support. Figure~\ref{fig:framework} illustrates these components through a financial-analysis example.

\textbf{Schema Layer.}\quad
The Schema Layer $\Gamma_t$ defines the fields of the four node families, the admissible Semantic Relation types, and the permitted reference patterns. Schema updates can therefore extend the ontology's representational capacity without changing its instantiated content.

\textbf{Tool Layer.}\quad
The Tool Layer $\mathcal{R}_t$ exposes the ontology through two MCP tools and a session manifest. The function $f_{\mathrm{browse}}(q,k,n)$ retrieves the top-$n$ semantic matches for query $q$ and kind $k$, while $f_{\mathrm{resolve}}(\mathcal{I},c)$ returns the requested records and their linked objects. The manifest provides compact source and usage information at session initialization. It is the only ontology content placed in the prompt, while detailed records are retrieved on demand.

\subsection{Evidence-Grounded Ontology Initialization}
Manually defining domain concepts, field mappings, linking paths, and semantic constraints for each data source requires substantial expert effort. The builder agent constructs an initial ontology from the training workload and raw sources without observing gold answers. The workload identifies semantics relevant to the agent, while executable probes verify their grounding in the underlying data.

\textbf{Workload-Guided Probing.}\quad
Given a training workload $\mathcal{W}$ and raw sources $\mathcal{D}$, the builder proposes
$\mathcal{C}=\mathrm{propose}(\mathcal{W})$
from recurrent entities, metrics, operations, and analytical conditions. For each candidate $c\in\mathcal{C}$, it issues
$\mathrm{probe}(c,\mathcal{D})$
to identify candidate fields and linking paths and to inspect their types, values, and semantic consistency.

\textbf{Evidence-Grounded Commitment.}\quad
Only candidates supported by their probe results are committed to the initial Content Layer:
\begin{equation}
\begin{aligned}
\mathcal{C}^{+}
&=
\left\{
c\in\mathcal{C}
\,\middle|\,
\mathrm{verify}\!\left(
\mathrm{probe}(c,\mathcal{D})
\right)=1
\right\},\\
\mathcal{S}_{0}
&=
\mathrm{construct}\!\left(
\mathcal{C}^{+},\mathcal{D};\Gamma_{0}
\right).
\end{aligned}
\end{equation}
Here, $\mathrm{verify}(\cdot)$ checks the declared type, filter, and value-distribution requirements. Verified candidates are instantiated under $\Gamma_0$, with their supporting records retained as Evidence. Together with the default Tool Layer $\mathcal{R}_0$, they form the initial state
$\mathcal{L}_0=(\mathcal{S}_0,\Gamma_0,\mathcal{R}_0)$.

\subsection{Trajectory-Grounded Ontology Evolution}

Data grounding alone does not ensure that an ontology suits a particular agent. EvoOntology therefore uses historical trajectories as behavioral evidence. Successful executions reveal effective semantic structures and access patterns, while unsuccessful ones expose missing, misleading, or poorly exposed components.

\textbf{Trajectory Attribution.}\quad
Given historical trajectories $\mathcal{T}_t$ and the current state $\mathcal{L}_t$, the evolution agent extracts recurrent signatures
$\Sigma_t=\mathrm{analyze}(\mathcal{T}_t,\mathcal{L}_t)$.
Each signature summarizes an interaction pattern, the ontology objects involved, and its observed outcomes. The agent assigns the signature to Content, Tool, or Schema through
$\alpha:\Sigma_t\rightarrow\{\mathsf{C},\mathsf{T},\mathsf{S}\}$
and states the expected behavioral effect of an update.

\textbf{Localized Intervention.}\quad
For an attributed signature $\sigma$, the agent proposes
$\mathcal{L}'_t=\mathrm{patch}(\mathcal{L}_t,\sigma,\alpha(\sigma))$.
Each candidate modifies one level only. Content interventions add, remove, or revise instantiated semantic objects in $\mathcal{S}_t$. Tool interventions modify existing tools or add and remove tools in $\mathcal{R}_t$ according to observed agent behavior. Schema interventions revise the object model in $\Gamma_t$. Multiple dependent Content objects may be updated together when they implement the same hypothesis.


\textbf{Backbone-Conditional Paired Validation.}\quad
For backbone $m$, let $\phi(\mathcal{L},\mathcal{V};m)$ denote the score of ontology state $\mathcal{L}$ on validation set $\mathcal{V}$. The candidate and its parent are evaluated on the same $\mathcal{V}$ with identical decoding and interaction budgets. The candidate is retained only when its improvement reaches margin $\tau$:
\begin{equation}
\mathcal{L}_{t+1}
=
\begin{cases}
\mathcal{L}'_t,
&
\phi(\mathcal{L}'_t,\mathcal{V};m)
-
\phi(\mathcal{L}_t,\mathcal{V};m)
\geq \tau,\\
\mathcal{L}_t,
&
\text{otherwise.}
\end{cases}
\end{equation}
The single-level difference isolates the attributed hypothesis while limiting regressions on the validation set. Rejected candidates are not deployed, and their signatures, interventions, and evaluation outcomes are logged to avoid repeated ineffective updates. All backbones evolve independently from the same initial state $\mathcal{L}_0$, allowing accepted updates to reflect backbone-specific interaction patterns.

\section{Experiments}
\subsection{Benchmarks}
We evaluate EvoOntology on three data-agent benchmarks with heterogeneous modalities and answer formats. All evaluations follow each benchmark’s official evaluation
protocol.

\textbf{Deep Data Research (DDR-Bench)}~\cite{ddr_bench} evaluates open-ended data research across heterogeneous sources. We evaluate on the 10-K scenario, and report \emph{Message-Wise} accuracy on per-turn interpretation, \emph{Trajectory-Wise} accuracy on full-history synthesis.

\textbf{InsightBench}~\cite{insightbench} is a business-analytics benchmark of business-intelligence flags, each paired with a CSV dataset and a ground-truth insight that an analyst should surface. We report the \emph{Insight} and \emph{Summary} scores.

\textbf{BIRD}~\cite{bird_bench} is a text-to-SQL benchmark on natural-language questions across real-world databases, evaluated under the official Oracle Knowledge setting. Follow-up benchmarks such as Spider~\cite{spider,spider2} extend the setting to multi-schema and enterprise workflows. The primary metric is \emph{Execution Accuracy} EX and the secondary is \emph{Valid Efficiency Score} VES.

\subsection{Experimental Setup}
\textbf{Backbones.}\quad We evaluate EvoOntology on six LLM backbones: GPT-5.5, GPT-5.6-sol, Claude-Sonnet-5, Claude-Opus-4.8, DeepSeek-V4-Flash, and Qwen3.5-Flash. For each backbone, all conditions use the same ReAct~\cite{react} scaffold, raw-data tools, decoding configuration, and interaction budget. Scoring follows each benchmark's standard evaluation protocol~\cite{bird_bench,insightbench,ddr_bench}.

\textbf{Baselines.}\quad We compare EvoOntology against two baselines under the same ReAct scaffold and backbone. \emph{Baseline} runs ReAct without any ontology layer, so the agent must rediscover the schema and the domain vocabulary at every task. \emph{Baseline + SL} prepends the builder-agent's semantic layer into the agent's context as a static prompt fragment~\cite{rsl_sql,e_sql,pet_sql,schema_prompt}.


\textbf{Reciprocal Two-Fold Evaluation.}\quad
We treat ontology construction and evolution as training-time workload adaptation, following held-out optimization protocols in prompt and agent adaptation~\cite{ape,yang2024large,xu2026adapting}. Each benchmark is divided into two disjoint folds, $A$ and $B$. In the $A\!\rightarrow\!B$ run, $70\%$ of $A$ is used for ontology construction, trajectory analysis, and candidate generation, and the remaining $30\%$ for paired validation. The selected ontology is frozen before testing on $B$. We then reverse the folds and report
\[
\mathrm{Score}
=
\frac{
\mathrm{Score}_{A\rightarrow B}
+
\mathrm{Score}_{B\rightarrow A}
}{2}.
\]
This reciprocal design follows two-fold split-and-swap evaluation~\cite{dietterich1998approximate,wang2026causaldetox}. All methods use the same fold assignment and deployment configuration. The same adaptation fold is used for ontology construction and updating across all relevant conditions. The held-out fold is accessed only for final evaluation after the ontology has been frozen, and its answers and evaluator feedback are never used for ontology construction, evolution, or candidate selection.

\subsection{Main Results}

\begin{table}[t]
\centering
\scriptsize
\setlength{\tabcolsep}{2pt}
\begin{tabular}{llccc}
\toprule
Method & Backbone
& \makecell{Msg-Wise\\(\%, $\uparrow$)}
& \makecell{Traj-Wise\\(\%, $\uparrow$)}
& \makecell{Overall\\(\%, $\uparrow$)} \\
\midrule
\multirow{8}{*}{\makecell[l]{Reported\\ReAct}}
& Claude-Sonnet-4.5 & 77.6 & 60.6 & 69.1 \\
& DeepSeek-V3.2     & 60.1 & 38.2 & 49.2 \\
& GLM-4.6           & 60.3 & 36.0 & 48.2 \\
& GPT-5.2           & 44.9 & 41.1 & 43.0 \\
& GPT-5-mini        & 46.8 & 37.1 & 42.0 \\
& Kimi-K2           & 51.1 & 30.8 & 40.1 \\
& GPT-5.1           & 37.1 & 44.3 & 40.7 \\
& Gemini-3-Flash    & 44.8 & 21.2 & 33.0 \\
\midrule

\multirow{6}{*}{\makecell[l]{Baseline\\(ReAct \\w/o Ontology)}}
& GPT-5.5           & 60.6 & 64.2 & 62.4 \\
& GPT-5.6-sol       & 64.0 & 68.5 & 66.3 \\
& Claude-Sonnet-5   & 74.3 & 72.5 & 73.4 \\
& Claude-Opus-4.8   & 74.0 & 73.0 & 73.5 \\
& DeepSeek-V4-Flash & 26.2 & 30.3 & 28.2 \\
& Qwen3.5-Flash     & 16.4 & 14.3 & 15.4 \\
\midrule

\multirow{6}{*}{\makecell[l]{Baseline + SL\\(ReAct +\\Semantic Layer)}}
& GPT-5.5           & 58.4 ($-$2.2) & 63.9 ($-$0.3)  & 61.2 ($-$1.2) \\
& GPT-5.6-sol       & 62.5 ($-$1.5) & 65.5 ($-$3.0)  & 64.0 ($-$2.3) \\
& Claude-Sonnet-5   & 65.6 ($-$8.7) & 57.5 ($-$15.0) & 61.5 ($-$11.9) \\
& Claude-Opus-4.8   & 65.9 ($-$8.1) & 71.4 ($-$1.6)  & 68.6 ($-$4.9) \\
& DeepSeek-V4-Flash & 28.8 (+2.6)   & 31.7 (+1.4)    & 30.2 (+2.0) \\
& Qwen3.5-Flash     & 14.8 ($-$1.6) & 13.3 ($-$1.0)  & 14.1 ($-$1.3) \\
\midrule

\multirow{6}{*}{\textbf{EvoOntology}}
& GPT-5.5           & 74.0 (+13.4) & 90.9 (+26.7) & 82.5 (+20.1) \\
& GPT-5.6-sol       & 78.2 (+14.2) & 93.5 (+25.0) & 85.9 (+19.6) \\
& Claude-Sonnet-5   & 78.4 (+4.1)  & 81.3 (+8.8)  & 79.9 (+6.5) \\
& Claude-Opus-4.8   & 78.0 (+4.0)  & 92.3 (+19.3) & 85.2 (+11.7) \\
& DeepSeek-V4-Flash & 37.5 (+11.4) & 52.3 (+22.0) & 44.9 (+16.7) \\
& Qwen3.5-Flash     & 21.1 (+4.7)  & 19.1 (+4.8)  & 20.1 (+4.8) \\
\bottomrule
\end{tabular}
\caption{Main results on the DDR-Bench 10-K scenario. Parentheses report the gain over the Baseline result.}
\label{tab:ddr}
\end{table}

\textbf{Capability on Multi-Source Data Research.}\quad Table~\ref{tab:ddr} reports DDR-Bench results across six LLM backbones. EvoOntology improves Trajectory-Wise accuracy on all six backbones, with an average gain of $+17.8$ points over Baseline. The improvement ranges from $+4.8$ on Qwen3.5-Flash to $+26.7$ on GPT-5.5, indicating that the ontology remains effective across backbones with substantially different baseline capabilities. In contrast, \emph{Baseline + SL}, which injects the semantic layer into the context as a static prompt, does not consistently improve over the un-mediated agent and even drops by $-15.0$ points on Claude-Sonnet-5. The gap between Baseline + SL and EvoOntology stems from how the layer is used: a static prompt fragment competes with the agent's other instructions and cannot be pruned per turn, whereas EvoOntology exposes the same content through MCP tools that the agent actively queries, retrieving only the terms and mappings relevant to the current step. We additionally compare against \emph{ReAct + Memory}~\cite{reflexion,voyager,self_refine}, which stores past trajectories as retrievable episodes. As shown in Table~\ref{tab:memory_compare}, memory-based persistence lifts Trajectory-Wise from $69.5$ to $75.8$ but remains $13.7$ points below EvoOntology, because episodic memory only replays what has been done and does not expose typed, composable structure.

\begin{table}[t]
\centering
\small
\setlength{\tabcolsep}{6pt}
\begin{tabular}{lcc}
\toprule
Method & Traj-Wise (\%, $\uparrow$) & $\Delta$ \\
\midrule
Baseline (ReAct)          & $69.5$        & --      \\
ReAct + Memory            & $75.8$        & $+6.3$  \\
\textbf{EvoOntology}      & \textbf{89.5} & $+20.0$ \\
\bottomrule
\end{tabular}
\caption{Comparison against a memory-based persistence baseline on DDR-Bench, averaged across the four backbones. ``ReAct + Memory'' stores past trajectories as retrievable episodes and injects the top-$k$ into the prompt.}
\label{tab:memory_compare}
\end{table}

\begin{table}[t]
\centering
\scriptsize
\setlength{\tabcolsep}{3pt}
\begin{tabular}{llccc}
\toprule
Method & Backbone
& \makecell{Insight\\(\%, $\uparrow$)}
& \makecell{Summary\\(\%, $\uparrow$)}
& \makecell{Overall\\(\%, $\uparrow$)} \\
\midrule
Pandas Agent
& GPT-4o
& 54.0 & 40.0 & 47.0 \\

AgentPoirot
& GPT-3.5-turbo
& 50.0 & 31.0 & 40.5 \\

AgentPoirot
& GPT-4-turbo
& 56.0 & 35.0 & 45.5 \\

AgentPoirot
& Llama-3-70B
& 52.0 & 33.0 & 42.5 \\

AgentPoirot
& GPT-4o
& 60.0 & 44.0 & 52.0 \\
\midrule

\multirow{6}{*}{\makecell[l]{Baseline\\(ReAct \\w/o Ontology)}}
& GPT-5.5           & 52.9 & 47.6 & 50.3 \\
& GPT-5.6-sol       & 51.6 & 49.4 & 50.5 \\
& Claude-Sonnet-5   & 53.3 & 51.3 & 52.3 \\
& Claude-Opus-4.8   & 54.9 & 49.9 & 52.4 \\
& DeepSeek-V4-Flash & 45.0 & 34.6 & 39.8 \\
& Qwen3.5-Flash     & 37.5 & 26.2 & 31.9 \\
\midrule

\multirow{6}{*}{\makecell[l]{Baseline + SL\\(ReAct +\\Semantic Layer)}}
& GPT-5.5
& 51.2 ($-$1.7) & 48.4 (+0.8) & 49.8 ($-$0.5) \\

& GPT-5.6-sol
& 51.3 ($-$0.3) & 50.8 (+1.4)   & 51.1 (+0.6) \\

& Claude-Sonnet-5
& 53.5 (+0.2)   & 48.0 ($-$3.3) & 50.8 ($-$1.6) \\

& Claude-Opus-4.8
& 54.3 ($-$0.6) & 50.2 (+0.3) & 52.3 ($-$0.1) \\

& DeepSeek-V4-Flash
& 47.0 (+2.0)   & 36.5 (+1.9)   & 41.8 (+2.0) \\

& Qwen3.5-Flash
& 39.0 (+1.5)   & 25.2 ($-$1.0) & 32.1 (+0.2) \\
\midrule

\multirow{6}{*}{\textbf{EvoOntology}}
& GPT-5.5
& 53.4 (+0.5) & 48.6 (+1.0) & 51.0 (+0.8) \\

& GPT-5.6-sol
& 53.2 (+1.6) & 50.9 (+1.5) & 52.1 (+1.6) \\

& Claude-Sonnet-5
& 54.4 (+1.1) & 51.5 (+0.2) & 53.0 (+0.7) \\

& Claude-Opus-4.8
& 55.8 (+0.9) & 50.5 (+0.6) & 53.2 (+0.8) \\

& DeepSeek-V4-Flash
& 49.2 (+4.2) & 42.6 (+8.0) & 45.9 (+6.1) \\

& Qwen3.5-Flash
& 39.3 (+1.8) & 27.6 (+1.4) & 33.4 (+1.6) \\
\bottomrule
\end{tabular}
\caption{Main results on InsightBench. Parentheses report the gain over the corresponding Baseline result.}
\label{tab:insightbench}
\end{table}

\textbf{Capability on Insight Mining.}\quad Table~\ref{tab:insightbench} reports InsightBench results across six backbones. EvoOntology improves Overall performance on every backbone, with a mean gain of $1.9$ points and the largest improvement on DeepSeek-V4-Flash ($+6.1$). The gains are smaller than DDR-Bench because Insight is graded on short reference-style findings and saturates once the answer aligns with the reference. \emph{Baseline + SL} shows mixed results, but drops by $-3.3$ on Claude-Sonnet-5 Summary, whereas EvoOntology improves both Insight and Summary on all backbones by exposing the same content through queryable tools instead of a static prompt.

\begin{table}[t]
\centering
\scriptsize
\setlength{\tabcolsep}{4pt}
\begin{tabular}{llcc}
\toprule
Method & Backbone
& EX (\%, $\uparrow$)
& VES (\%, $\uparrow$) \\
\midrule
GPT-4   & GPT-4  & 46.4 & --   \\
DIN-SQL & GPT-4  & 50.7 & 58.8 \\
DAIL-SQL & GPT-4 & 54.8 & 56.1 \\
TA-SQL  & GPT-4  & 56.2 & --   \\
MAC-SQL & GPT-4  & 57.6 & 58.8 \\
MCS-SQL & GPT-4  & 63.4 & --   \\
CHESS   & GPT-4o & 65.0 & 62.8 \\
\midrule

\multirow{6}{*}{\makecell[l]{Baseline\\(ReAct w/o Ontology)}}
& GPT-5.5           & 61.5 & 63.4 \\
& GPT-5.6-sol       & 63.5 & 65.6 \\
& Claude-Sonnet-5   & 61.9 & 63.7 \\
& Claude-Opus-4.8   & 67.5 & 69.6 \\
& DeepSeek-V4-Flash & 33.1 & 36.4 \\
& Qwen3.5-Flash     & 46.5 & 47.9 \\
\midrule

\multirow{6}{*}{\makecell[l]{Baseline + SL\\(ReAct +\\Semantic Layer)}}
& GPT-5.5           & 55.9 ($-$5.6) & 67.7 (+4.3) \\
& GPT-5.6-sol       & 63.0 ($-$0.5) & 68.9 (+3.3) \\
& Claude-Sonnet-5   & 60.8 ($-$1.1) & 65.8 (+2.1) \\
& Claude-Opus-4.8   & 66.2 ($-$1.3) & 75.0 (+5.4) \\
& DeepSeek-V4-Flash & 36.3 (+3.2)   & 37.2 (+0.7) \\
& Qwen3.5-Flash     & 48.0 (+1.5)   & 51.9 (+4.0) \\
\midrule

\multirow{6}{*}{\textbf{EvoOntology}}
& GPT-5.5           & 68.9 (+7.4)  & 71.1 (+7.7) \\
& GPT-5.6-sol       & 70.7 (+7.2)  & 73.0 (+7.4) \\
& Claude-Sonnet-5   & 71.8 (+9.9)  & 74.1 (+10.4) \\
& Claude-Opus-4.8   & 78.3 (+10.8) & 80.5 (+10.9) \\
& DeepSeek-V4-Flash & 39.4 (+6.4)  & 44.1 (+7.6) \\
& Qwen3.5-Flash     & 49.1 (+2.5)  & 55.2 (+7.3) \\
\bottomrule
\end{tabular}
\caption{Main results on BIRD under Oracle Knowledge. VES is reported on a $0$--$100$ scale. Parentheses report the gain over the corresponding Baseline result.}
\label{tab:bird}
\end{table}

\textbf{Capability on Data Retrieval.}\quad Table~\ref{tab:bird} reports BIRD results across six backbones under Oracle Knowledge. EvoOntology improves both EX and VES for every backbone, with average gains of $7.4$ and $8.6$ points. The consistent gains across both metrics indicate that the ontology improves query correctness as well as execution efficiency. \emph{Baseline + SL} shows a mixed pattern: EX drops by up to $-5.6$ (GPT-5.5), while VES rises across all six backbones, suggesting better efficiency-aware performance despite mixed effects on execution accuracy. Once the same content is exposed through MCP tools that the agent actively queries and refined by the evolution loop, EvoOntology recovers the EX gains and yields a stable per-backbone improvement over both baselines and prior text-to-SQL systems~\cite{din_sql,mac_sql,chess}.

\subsection{Effect of Ontology Layer}

\begin{figure}[t]
\centering
\includegraphics[width=\columnwidth]{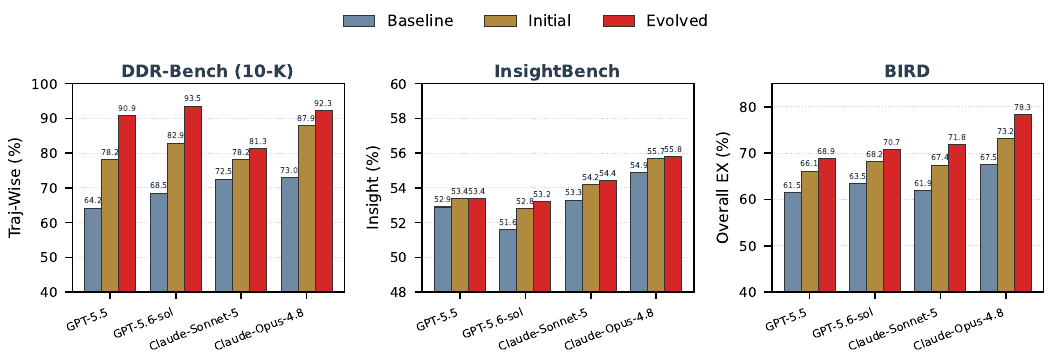}
\caption{Primary metric on the three benchmarks under three conditions: \emph{Baseline} , \emph{Initial}, and \emph{Evolved} (EvoOntology).}
\label{fig:progression}
\end{figure}

To separate the contribution of the builder-constructed ontology from the additional gain brought by self-evolution, we compare three settings: \emph{Baseline}, \emph{Initial}, and \emph{Evolved}. \emph{Baseline} uses no ontology layer, \emph{Initial} uses the ontology constructed by the builder agent before evolution, and \emph{Evolved} uses the final ontology after self-evolution. Figure~\ref{fig:progression} reports the performance of each backbone under the three settings. To summarize the overall trend, we average the primary-metric scores across the four backbones for each benchmark and setting and compare the resulting means. 

The initial ontology establishes a strong improvement over the no-ontology baseline, while self-evolution consistently extends this gain across all three benchmarks. On DDR-Bench, the mean Trajectory-Wise score increases by $12.3$ percentage points from \emph{Baseline} to \emph{Initial}, followed by a further improvement of $7.7$ percentage points from \emph{Initial} to \emph{Evolved}. On InsightBench, the mean Insight score first increases by $0.8$ points and then gains another $0.2$ points through evolution. On BIRD, the mean EX score improves by $5.1$ percentage points with the initial ontology and by a further $3.7$ percentage points after evolution. These results show that the builder-constructed ontology provides an effective starting point, whereas the self-evolution loop is essential for realizing the full performance gain and consistently improves the ontology beyond its initial state.

\section{Analyses}
To better understand the source and behavior of EvoOntology's advantage, we conduct a series of in-depth analyses. Unless otherwise stated, all analyses in this section are conducted on DDR-Bench across the four backbones (GPT-5.5, GPT-5.6-sol, Claude-Sonnet-5, Claude-Opus-4.8).

\subsection{Effect of Iterative Evolution}

\begin{figure}[t]
\centering
\includegraphics[width=\columnwidth]{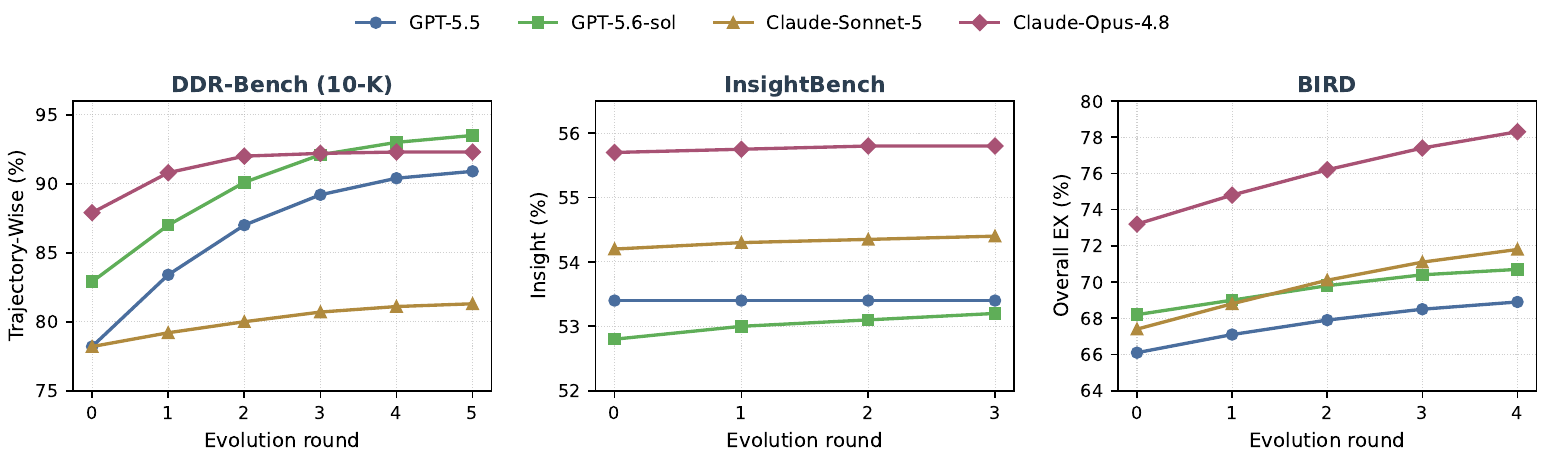}
\caption{Primary metric across accepted evolution rounds on the three benchmarks: Traj-Wise on DDR-Bench, Insight on InsightBench, and EX on BIRD.}
\label{fig:evo_trajectory}
\end{figure}

To evaluate whether the observed gain accumulates through many small edits and does not collapse into a single round, we plot the deployed agent's primary score across the sequence of accepted evolution rounds on DDR-Bench. Each round corresponds to one candidate that passed the paired gate, and the parent line traces the score of the ontology version that would remain if no more rounds were run. As shown in Figure~\ref{fig:evo_trajectory}, all four backbones improve monotonically from Initial through the accepted rounds, with GPT-5.6-sol reaching $93.5$ Traj-Wise after five accepted rounds and Claude-Opus-4.8 reaching $92.3$ after four. Notably, the trajectories flatten by the last two rounds, which is consistent with the failure signatures becoming rarer once the ontology covers the recurrent cross-filing concepts. The results show that the gains reported in Table~\ref{tab:ddr} are the outcome of a converging refinement and not a single fortunate patch, which validates the design of the four-step evolution loop.

\subsection{Ablation Study on Evolution Loop}

The relative contribution of the four steps in the evolution loop (diagnose, attribute, patch, gate) is assessed by disabling each step in turn and comparing the resulting final Evolved score on DDR-Bench, averaged across the four backbones.The disabled variant of each step is: \emph{w/o Diagnose} skips the failure-trace clustering step and asks the evolution agent to propose an edit from a random sample of recent traces; \emph{w/o Attribution} drops the level tag and lets the agent commit an edit at any level without stating a hypothesis; \emph{w/o Patch stage} replaces the typed, hypothesis-conditioned edit with a free-form ontology rewrite that the evolution agent produces directly from the diagnosis; \emph{w/o Gate} accepts every candidate patch. As shown in Table~\ref{tab:evo_ablation}, removing the gate causes the largest drop ($-11.2$ Traj-Wise), because unfiltered candidates admit regressions that the next round cannot always undo. Removing the attribution step drops by $-6.3$, because without a level tag the loop tends to make content edits when the failure is a manifest problem, and vice versa. Removing the diagnose step drops by $-4.8$, and replacing the typed patch with a free-form rewrite drops by $-1.7$. The results show that the gate and attribution are the two load-bearing pieces, which validates the design of an evolution loop that is more selective than iterative.

\begin{table}[t]
\centering
\small
\setlength{\tabcolsep}{6pt}
\begin{tabular}{lcc}
\toprule
Variant & Traj-Wise (\%, $\uparrow$) & $\Delta$ \\
\midrule
Full loop                & \textbf{89.5} & --      \\
\quad w/o Gate           & $78.3$        & $-11.2$ \\
\quad w/o Attribution    & $83.2$        & $-6.3$  \\
\quad w/o Diagnose       & $84.7$        & $-4.8$  \\
\quad w/o Patch (free-form) & $87.8$        & $-1.7$  \\
\bottomrule
\end{tabular}
\caption{Ablation on the four steps of the evolution loop, averaged across four backbones.}
\label{tab:evo_ablation}
\end{table}



\textbf{Three-Level Evolution.}\quad Beyond removing individual steps, we evaluate whether the three editable levels (Content / Tool / Schema) are jointly required by restricting the evolution loop to a single level at a time and comparing against Initial on DDR-Bench, averaged across the four backbones. As shown in Table~\ref{tab:three_layer_ablation}, single-level evolution yields limited gains and may even hurt performance: Tool-only achieves only a marginal improvement ($+0.9$), while Content-only and Schema-only decrease performance by $3.6$ and $8.7$ points, respectively. In contrast, full three-level evolution achieves a substantial improvement ($+7.7$) over Initial, outperforming the best single-level variant by $6.8$ points. These results demonstrate the effectiveness of our three-level evolution design, showing that jointly evolving the three levels is more effective than optimizing any single level in isolation.

\begin{table}[t]
\centering
\small
\setlength{\tabcolsep}{6pt}
\begin{tabular}{lcc}
\toprule
Variant & Traj-Wise (\%, $\uparrow$) & $\Delta$ \\
\midrule
Initial & $81.8$ & -- \\
\quad Content-only evolution  & $78.2$ & $-3.6$ \\
\quad Tool-only evolution     & $82.7$ & $+0.9$ \\
\quad Schema-only evolution   & $73.1$ & $-8.7$ \\
Full three-level evolution   & \textbf{89.5} & $+7.7$ \\
\bottomrule
\end{tabular}
\caption{Ablation on the three editable levels of the evolution loop on DDR-Bench, averaged across four backbones.}
\label{tab:three_layer_ablation}
\end{table}

\subsection{Ablation Study on Ontology Structure}

We mask each removable object family from the final \emph{Evolved} ontology on DDR-Bench and report the average performance across four backbones. As shown in Table~\ref{tab:family_ablation}, masking Mappings causes the largest drop ($-13.4$ Traj-Wise), which is consistent with the role of Mappings as the only object that grounds a Term to concrete columns and join paths. Masking Evidence drops by $-8.7$, because without a probe query the agent cannot verify a candidate SQL fragment against the underlying value distribution. Masking Constraints and Relations produces smaller drops ($-3.5$ and $-2.1$), and Terms cannot be masked in isolation as every other family references them. These findings identify Mappings and Evidence as the two load-bearing families, which validates our decision to require every committed entry to be anchored in a probe query and not a natural-language description alone.

\begin{table}[t]
\centering
\small
\setlength{\tabcolsep}{6pt}
\begin{tabular}{lcc}
\toprule
Variant & Traj-Wise (\%, $\uparrow$) & $\Delta$ \\
\midrule
Full EvoOntology            & \textbf{89.5} & --      \\
\quad w/o Mappings    & $76.1$        & $-13.4$ \\
\quad w/o Evidence    & $80.8$        & $-8.7$  \\
\quad w/o Constraints & $86.0$        & $-3.5$  \\
\quad w/o Relations   & $87.4$        & $-2.1$  \\
\bottomrule
\end{tabular}
\caption{Ablation on the five object families of the ontology content layer on DDR-Bench, averaged across four backbones. Terms cannot be masked in isolation and are omitted.}
\label{tab:family_ablation}
\end{table}

\subsection{Divergence across Backbones}

\begin{figure}[t]
\centering
\begin{subfigure}[t]{0.48\columnwidth}
\centering
\includegraphics[width=\linewidth]{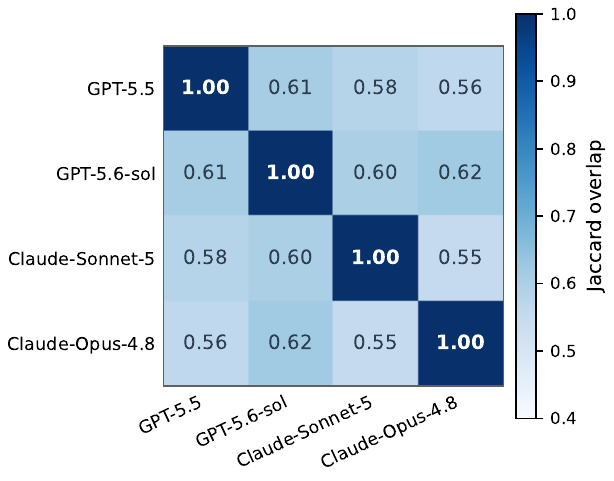}
\caption{Pairwise Jaccard overlap of accepted Term identifiers between the evolved stores of the four backbones.}
\label{fig:store_divergence}
\end{subfigure}
\hfill
\begin{subfigure}[t]{0.48\columnwidth}
\centering
\includegraphics[width=\linewidth]{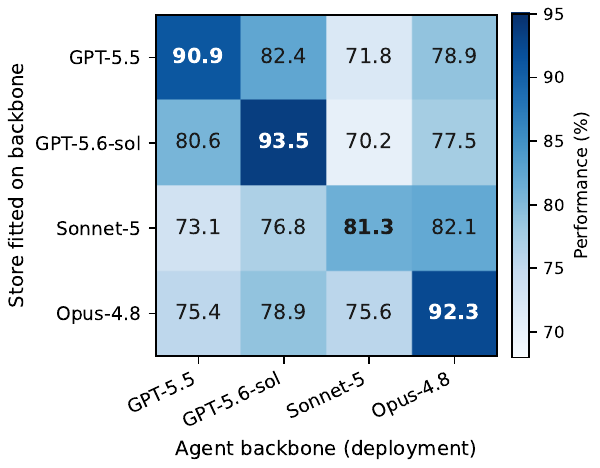}
\caption{Cross-backbone transfer of the evolved store: each row is fitted on one backbone and served to every backbone (columns).}
\label{fig:transfer_matrix}
\end{subfigure}
\caption{Generalization of the evolved ontology store across backbones on DDR-Bench.}
\label{fig:divergence}
\end{figure}

We investigate whether different backbones converge to similar ontologies or develop distinct ones by comparing the pairwise Jaccard overlap of their accepted Term-identifier sets on DDR-Bench. As shown in Figure~\ref{fig:store_divergence}, no pair exceeds $0.62$ overlap, and the two Claude backbones share less with each other ($0.55$) than the two GPT backbones do ($0.61$). The accepted edits also differ across backbones. For example, Claude-Opus-4.8 retains more detailed manifest variants than Claude-Sonnet-5, while GPT-5.5 introduces short SQL fragment libraries under Evidence that do not appear in the Claude-Opus-4.8 ontology. However, identifier overlap alone cannot determine semantic equivalence, since different identifiers may encode similar concepts. We further evaluate cross-backbone transfer by applying each evolved store to all four backbones and measuring Traj-Wise performance on DDR-Bench. As shown in Figure~\ref{fig:transfer_matrix}, the diagonal is uniformly the highest entry of its column, and every off-diagonal drops by at least 5.7 points relative to the same-backbone store; the average column drop from diagonal to off-diagonal ranges from -8.8 (Sonnet-5) to -14.5 (GPT-5.5). These results show that different backbones produce different evolved ontology stores from the same initialization. The cross-backbone transfer results further indicate that backbone-specific evolution is beneficial.

\section{Conclusion}
In this paper, we introduce \emph{EvoOntology}, an interactive ontology layer that is automatically constructed and self-evolving for data agents. EvoOntology encapsulates the ontology as an MCP server that the agent actively queries at runtime, and refines it through attribution-guided typed edits admitted only after a backbone-conditional paired evaluation gate. Experiments on benchmarks and six LLM backbones, EvoOntology consistently outperforms both ReAct baselines and traditional semantic-layer baselines, offering an effective solution for helping data agents understand heterogeneous data.

\bibliography{aaai2027}

\clearpage
\appendix
\section{Content-Layer Growth across Evolution Rounds}
\label{sec:store_growth}

\begin{figure}[t]
    \centering
    \includegraphics[width=\columnwidth]{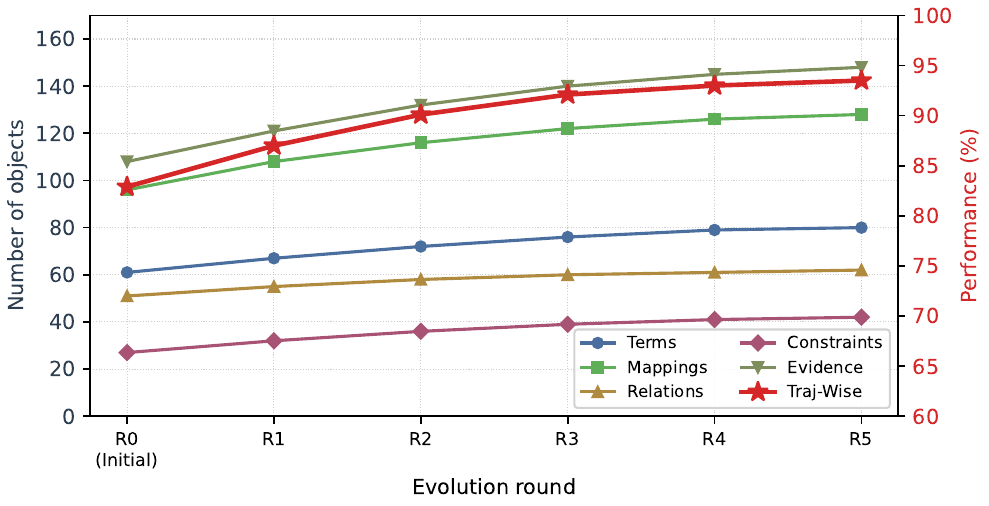}
    \caption{Growth of the Content Layer across accepted evolution
    rounds on DDR-Bench under GPT-5.6-sol. The curves report the four
    node families and instantiated Semantic Relations. The right axis
    reports Trajectory-Wise performance.}
    \label{fig:store_growth}
\end{figure}

To examine whether iterative evolution causes uncontrolled expansion of
the ontology content, we track its instantiated elements across the
accepted evolution rounds on DDR-Bench, using GPT-5.6-sol as a
representative backbone. The tracked elements comprise the four node
families, \emph{Terms}, \emph{Mappings}, \emph{Constraints}, and
\emph{Evidence}, together with instantiated \emph{Semantic Relations}.As shown in Figure~\ref{fig:store_growth}, most content growth occurs in
the first three rounds. The number of \emph{Terms} increases from $61$
in the \emph{Initial} ontology to $80$ after five accepted rounds,
while the per-round growth of every tracked element falls below $5\%$
after round three. The content-size curves then flatten together with
\emph{Trajectory-Wise} performance. Content expansion is therefore
concentrated in the early rounds, when the evolution loop addresses
recurrent semantic gaps, and stabilizes once these gaps have been
covered.

\section{Cost of the Ontology Layer}
\label{sec:ontology_cost}

The ontology layer introduces a compact manifest into the agent's
initial context and retrieves detailed semantic records through MCP
tools. We measure its computational cost using the average input and
output tokens per turn, the number of turns per task, and the resulting
total tokens per task on DDR-Bench.Table~\ref{tab:cost} shows that the \emph{Initial} ontology increases
average input tokens per turn from $3.2$K to $4.1$K because of the
manifest and retrieved semantics. At the same time, the average
trajectory shortens from $14.6$ to $11.2$ turns, reducing the total cost
from $52.6$K to $50.4$K tokens per task. The \emph{Evolved} ontology
further reduces the trajectory to $8.4$ turns and the total cost to
$42.0$K tokens, which is approximately $20\%$ below the
\emph{Baseline}. Over the same comparison, \emph{Trajectory-Wise}
performance rises from $69.5$ to $89.5$.The ontology layer therefore adds modest per-turn context while reducing repeated schema discovery over the full trajectory. Evolution strengthens this effect by improving how the agent discovers and grounds relevant semantics.

\begin{table}[t]
    \centering
    \small
    \setlength{\tabcolsep}{5pt}
    \begin{tabular}{lccc}
        \toprule
        Metric & \emph{Baseline} & \emph{Initial} & \emph{Evolved} \\
        \midrule
        Input tokens / turn (K)  & 3.2  & 4.1  & 4.6 \\
        Output tokens / turn (K) & 0.4  & 0.4  & 0.4 \\
        Turns / task             & 14.6 & 11.2 & \textbf{8.4} \\
        Total tokens / task (K)  & 52.6 & 50.4 & \textbf{42.0} \\
        \emph{Traj-Wise} (\%, $\uparrow$)
                                 & 69.5 & 81.8 & \textbf{89.5} \\
        \bottomrule
    \end{tabular}
    \caption{Cost of the ontology layer on DDR-Bench, averaged across
    the four-backbone analysis subset.}
    \label{tab:cost}
\end{table}

\section{Attribution across Editable Levels}
\label{sec:level_attribution}

\begin{figure}[t]
    \centering
    \includegraphics[width=\columnwidth]{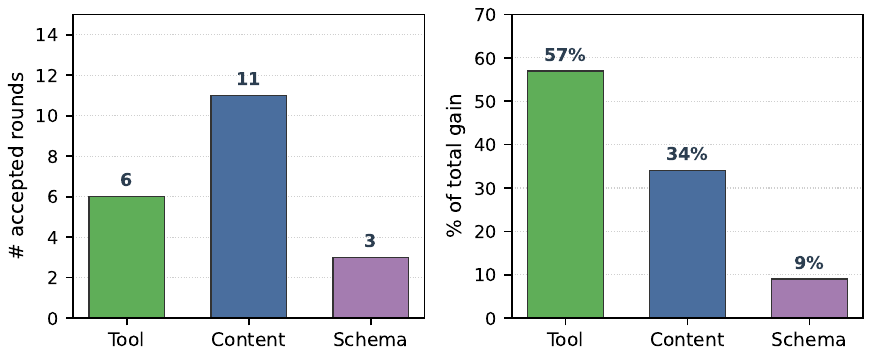}
    \caption{Distribution of the accepted evolution gain across
    Content, Tool, and Schema edits on DDR-Bench, aggregated over the
    four-backbone analysis subset.}
    \label{fig:layer_attribution}
\end{figure}

We next examine how the accepted evolution gain is distributed across
the three editable levels. Each accepted round is grouped by its
attribution tag, and the paired-evaluation improvement contributed by
each group is aggregated across the four backbones. As shown in Figure~\ref{fig:layer_attribution}, Tool-level edits account for $57\%$ of the cumulative gain across six accepted rounds. These edits mainly improve how existing ontology content is exposed through the manifest and MCP tools. Content-level edits contribute $34\%$
across eleven accepted rounds by adding or refining \emph{Terms},
\emph{Mappings}, \emph{Constraints}, \emph{Evidence}, and
\emph{Semantic Relations} identified from interaction trajectories.
Schema-level edits contribute the remaining $9\%$ across three accepted
rounds by changing the representational structure of the ontology. Content edits are more frequent, while Tool edits contribute the largest
share of the accumulated gain. Schema edits are less common but address
limitations that cannot be resolved by modifying instantiated content
alone. This distribution is consistent with the three levels serving
distinct and complementary roles during evolution.

\begin{figure*}[t]
    \centering
    \includegraphics[width=0.98\textwidth]{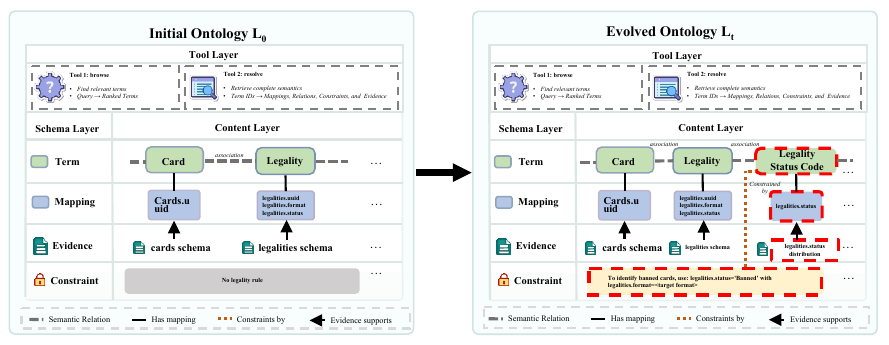}
    \caption{Evolution of the ontology for a card-legality task.
    The Initial state contains general Card and Legality semantics but
    no explicit interpretation of legality status. The accepted patch
    adds a Legality Status Code Term, its Mapping and Evidence, and a
    Constraint that relates the status value to the requested format.
    Red dashed boxes mark the added or refined objects.}
    \label{fig:appendix_case_study}
\end{figure*}

\section{Case Study: Evolution of Card-Legality Semantics}
\label{sec:card_case}
Figure~\ref{fig:appendix_case_study} presents a representative text-to-SQL case in which the agent must identify cards that are banned in a target game format. The case illustrates how a localized Content-level update extends the ontology without rewriting its existing Tool or Schema layers.

\textbf{Initial state.}
The \emph{Initial} ontology $\mathcal{L}_0$ contains the Terms
\emph{Card} and \emph{Legality}, together with an association between
them. The \emph{Card} Term is grounded to \texttt{Cards.uuid}, while
the \emph{Legality} Term is grounded to
\texttt{legalities.uuid}, \texttt{legalities.format}, and
\texttt{legalities.status}. Schema observations for the two tables are
retained as Evidence.Although these objects allow the agent to locate the relevant table,
the ontology does not explain how the values of
\texttt{legalities.status} should be interpreted. It also does not make
explicit that legality status is defined relative to a particular game
format. The agent must therefore rediscover these semantics from raw
values during execution.

\textbf{Attributed limitation.}
The evolution agent attributes this limitation to the Content Layer.
The existing \texttt{browse} and \texttt{resolve} tools can already
retrieve the relevant objects, and the Schema Layer can represent the
required knowledge. The missing component is a reusable semantic
description of the status field and its applicability condition.

\textbf{Localized intervention.}
The Candidate adds a new Term, \emph{Legality Status Code}, and grounds
it to \texttt{legalities.status}. An Evidence object records the
observed distribution of the status values. A Constraint then states
that identifying banned cards requires both
\texttt{legalities.status = 'Banned'} and
\texttt{legalities.format = target\_format}. The existing
\emph{Card} and \emph{Legality} objects remain unchanged, and the
Candidate introduces no Tool- or Schema-level modification.After passing paired validation, the Candidate becomes part of the
\emph{Evolved} ontology $\mathcal{L}_t$.

\textbf{Effect on agent interaction.}
With the \emph{Evolved} ontology, \texttt{browse} can surface
\emph{Legality Status Code} for queries involving banned or legal
cards. The agent can then use \texttt{resolve} to obtain the physical
Mapping, the supporting Evidence, and the format-dependent Constraint.
Native SQL execution remains responsible for applying the filter and
verifying the returned records.The case shows that evolution can correct a specific semantic gap by
adding a small connected set of objects. The ontology retains its
existing structure and interface while providing the agent with the
missing interpretation required for the task.

\end{document}